\documentclass[lettersize,journal]{IEEEtran}
\usepackage{amsmath,amsfonts}
\usepackage{algorithmic}
\usepackage{algorithm}
\usepackage{array}
\usepackage{subfigure}
\usepackage{textcomp}
\usepackage{stfloats}
\usepackage{url}
\usepackage{verbatim}
\usepackage{graphicx}
\usepackage{cite}
\usepackage{xcolor}
\usepackage{subfiles}
\usepackage{multirow}
\usepackage{amsmath}
\usepackage{subcaption}
\usepackage{soul}
\usepackage{mathtools}
\begin{document}

\title{Lane Change Intention Prediction of two distinct Populations using a Transformer based Neural Network}

\author{Francesco De Cristofaro, Cornelia Lex, Jia Hu, Arno Eichberger
    \thanks{This work has been submitted to the IEEE for  possible publication. Copyright may be transferred without notice, after which this version may no longer be accessible.}}

\markboth{}%
{Shell \MakeLowercase{\textit{et al.}}: A Sample Article Using IEEEtran.cls for IEEE Journals}

\IEEEpubid{}

\maketitle

\begin{abstract}
In complex traffic scenarios, intention prediction of surrounding vehicles can improve the strategy of automated driving functions. Existing work on intention prediction is often trained on datasets of single regions or countries. In this article, a transformer network for lane change intention prediction was trained to predict whether target vehicles perform a left lane change, right lane change or keep their lane. Features inputs used were vehicle positions and distances – each for longitudinal and lateral direction. Before being inputted, these features were converted to Frenet coordinates. Two different datasets from leveLXData collected on highways were used: one from German highways and one from Hong Kong highways. Through cross-dataset evaluation, we show that the accuracy values drop to $71.94\%$, compared to $85.44\%$ when doing dataset-specific training. When training on both datasets, accuracy levels up to $86.84\%$ were achieved.
\end{abstract}

\begin{IEEEkeywords}
Motion prediction, intention prediction, lane
change prediction, motion planning, decision making, automated driving, autonomous driving, artificial intelligence.
\end{IEEEkeywords}

\section{Introduction}\label{sec:introduction}
With the goal of increasing the safety and efficiency of the driving experience, vehicle manufacturers and governments have started to invest in assisted and automated driving technologies in the recent years with the possible end goal of achieving the full automation of passenger and commercial vehicles. The prediction of human's drivers' next maneuver could impact both safety and efficiency and has the potential of impacting the future of the car industry by improving the path planning capabilities of autonomous vehicles.

Given the importance of this topic, it is unsurprising that a lot of research has been already done in this direction. In \cite{ref1} and \cite{ref8} authors apply SVMs (Support Vector Machines) to trajectories collected through test vehicles to predict lane keeping (LK), left lane changes (LLCs) and right lane changes (RLCs). In \cite{ref10}, an SVM is used again, but, in this case, to predict LK and lane changes (LCs) without distincion for the direction of the maneuver. In \cite{ref9}, \cite{ref13} and \cite{ref86} authors explore the possibility of using a well established method for predicting LCs, RNN (recursive neural network), applying it on one of the most widely studied datasets of naturalistic trajectories, NGSIM (see \cite{ref_NGSIM}). Finally, in \cite{ref16}, authors test a transformer based network trained to predict LK, LLCs and RLCs on both American (NGSIM) and German (highD, see \cite{ref_highD}) data. 

While most authors approached the problem by selecting a suitable dataset of naturalistic trajectories to test their methods, no research was done regarding the possibility of training a method on a dataset to then deploy it in a region different to the one in which the dataset was collected and even \cite{ref16}, although using data from two different countries, does not provide cross test-training results or analysis of the population dependent differences in the performances on the two datasets.

In this paper, we try to analyze these aspects by using the exiD dataset \cite{exiD} and the Hong Kong dataset, both collected by leveLXData \cite{levelXdata}, which contain naturalistic trajectories recorded on highways/freeways, to train transformer networks to predict lane change maneuver within an upcoming time interval. We will in particular concentrate on the differences in performances between transformers trained on different (combinations of) datasets and we will try to answer two questions: 

\begin{itemize}
    \item Is it possible to train a transformer on a population and deploy the trained model in a different population?
    \item Does training a transformer network on two populations result in a loss of performance when testing the transformer on a singular population?
\end{itemize} 

The paper is structured as follows: in Section \ref{sec:data} the problem is described and the exiD and Hongkong datasets are presented and briefly discussed in addition to an explanation of the data processing. In Section \ref{sec:methodology} transformers are introduced and the task of designing them is described. In Section \ref{sec:results} the experiments are explained and the results of the prediction task are presented. In Section \ref{sec:results} the results are discussed and interpreted. Finally, Section \ref{sec:conclusion} contains our final comments and recommendations for future developments of the research.

The work presented in this article is a continuation of the work presented \cite{lc_pred}. Therein, different methods were compared to predict lane change intention predictions. In this paper we take one of the methods, a transformer, and we test it on different datasets collected from different regions. For this reason, parts of this paper, images and formulas might resemble or might be taken from the previous work. The results obtained in this paper are, though, completely novel and have not been presented in earlier publications.

\section{Problem Definition and Input Data}\label{sec:data}
In this work, both the exiD dataset \cite{exiD} and the Hongkong dataset will be used. The exiD dataset is a dataset of naturalistic driving trajectory collected by leveLXData on German highways using drones at a sampling frequency of $25$ Hz \cite{exiD}. The dataset includes $16$ hours of measurement data for a total of  $69 172$ vehicle trajectories recorded on $7$ different locations (roads). The Hongkong dataset is a similarly structured dataset also collected by leveLXData on Hong Kong's, China, highways and freeways using drones at a sampling frequency of $30$ Hz. The dataset includes $13.8$ hours of measurement data for a total of  $99 842$ vehicle trajectories recorded on $5$ different locations (roads). Before proceeding with the processing and labeling of the dataset, it is important to understand which scenario is considered in this work, which problem is tackled and how data is used to solve it. In this section these themes will be dealt with and the data preparation will be explained in detail.

\subsection{Scenario Definition}
\begin{figure}
    \centering
    \includegraphics[width=0.9\linewidth]{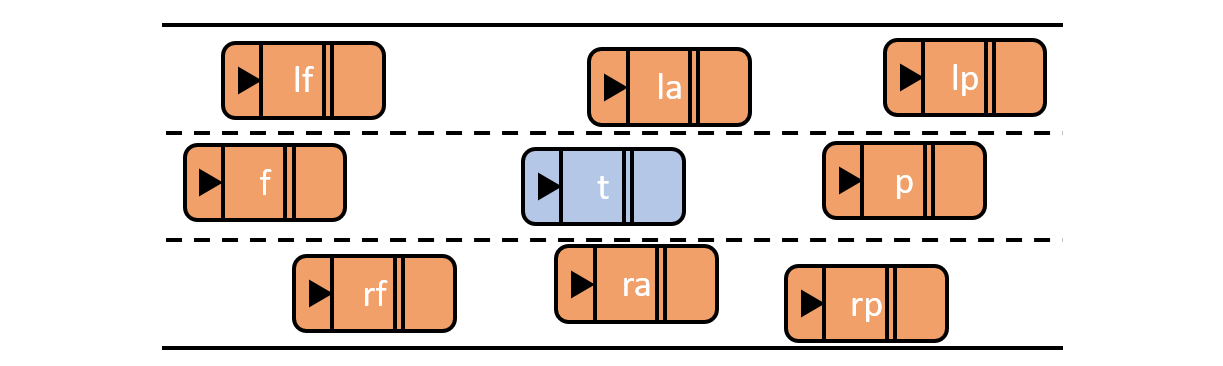}
    \caption{Scenario considered in this work. The target (t) vehicle is surrounded by the right following vehicle (rf), the right alongside vehicle (ra), the right preceding vehicle (rp), the following alongside vehicle (fa), the preceding vehicle (p), the left following vehicle (lf), the left alongside vehicle (la) and the left preceding vehicle (lp). Figure taken from \cite{lc_pred}, licensed by CC BY 4.0 (see \cite{ccby4.0}).}
    \label{fig:scenario}
\end{figure}

This work focuses on highway scenarios. The objective is to predict the behavior of a single vehicle called target vehicle. In doing so its surrounding environment will also be taken in consideration, see Fig. \ref{fig:scenario}. A maximum of eight surrounding vehicles will be taken in consideration. Both datasets (exiD and Hongkong) include a small number of frames for which two vehicles are listed as alongside on the same lane. This happens due to how the data was processed. Given the small amount of data which these cases make up, they were not taken in consideration for prediction (the relative target vehicles are still used as surrounding vehicles for other target vehicles though). Surrounding vehicles driving on an on-ramp or off-ramp are considered only if their lateral distance to the target vehicles is smaller or equal to $6.0$ m to account for complex road structures. 

\subsection{Problem Definition}
The goal is to predict if the target vehicle will perform a left lane change maneuver (LLC) or left right change maneuver (RLC)  within the next $\Delta t_{p,\textnormal{MAX}}$ seconds (maximum prediction time) or if it will perform a lane keeping maneuver (LK), similarly to \cite{lc_pred}. To select an intermediate case between efficency and safety, $\Delta t_{p,\textnormal{MAX}}$ was set to $4$s. The problem is hence a multi-classification problem with three output classes. To predict a LC, the last $\Delta t_o$ seconds (observation window) of the trajectory of the target vehicle (the vehicle on which the prediction will be 
made) are observed and used as an input for the prediction algorithm. In particular, in this work $\Delta t_o=1$s 

In order to train a machine learning (ML) model to be able to perform such prediction, it is necessary to prepare a number of trajectories of uniform length extracted from the exiD and Hongkong datasets, label them accordingly if they precede a LK, a LLC or a RLC and use them to train and test said method.

\subsection{Coordinates conversion from Cartesian to Frenet}
Unlike highD dataset \cite{ref_highD}, a highly used dataset for training lane change intention prediction methods, both exiD dataset and Hongkong dataset do not include solely straight roads. They include both straight and curved roads with on-ramps and off-ramps. As it will be later explained, input features of the proposed transformer are longitudinal and lateral positions and longitudinal and lateral distances to surrounding vehicles. The coordinate systems used is a local Cartesian coordinate system $(x, y)$, one for each road. To ease the calculation of the input features, a transformation to Frenet coordinates is desirable since it makes the calculation of longitudinal and lateral quantities immediate. The aim is then to pass from $x$, $y$ to $s$, $l$ which respectively are the longitudinal position and lateral position in Frenet coordinates. Out of simplicity the conversion will be presented as if there was a single road and a single driving direction in the datasets but the reader should bear in mind that each road and driving direction included in the datasets necessitates a specific coordinate conversion since each road and driving direction results in a difference reference path.

\begin{figure}
    \centering
    \includegraphics[width=1\linewidth]{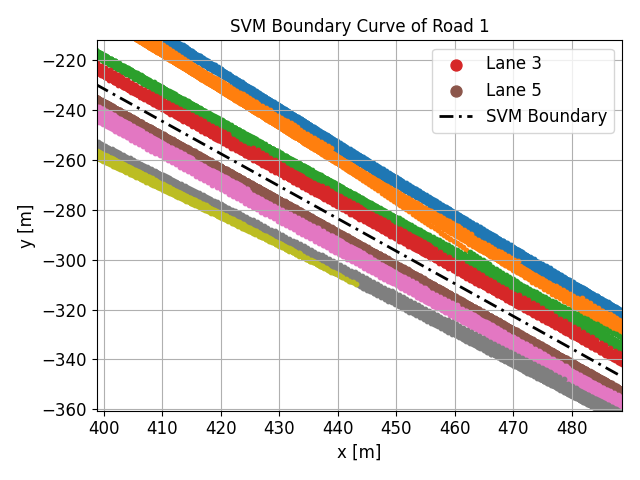}
    \caption{Boundary line resulting from the application of SVM to road 1 of the exiD dataset. The scattered points are trajectory points to which a color is assigned depending on the lane they are occupying.}
    \label{fig:SVM}
\end{figure}

\begin{figure}
    \centering
    \includegraphics[width=1\linewidth]{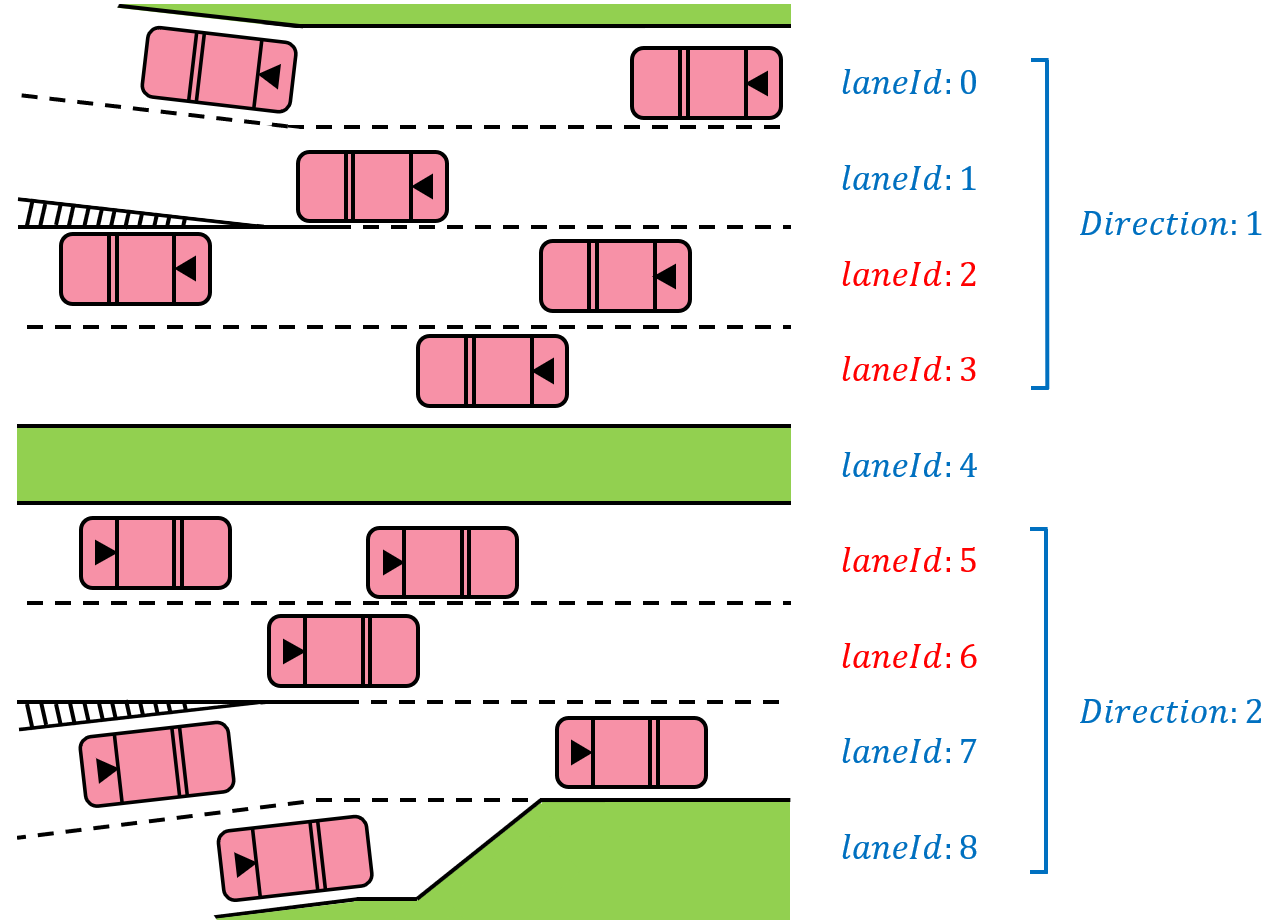}
    \caption{Lane numeration as defined for a section of road 1 in the exiD dataset. In red, the lanes which do not contain on-ramps or off-ramps. Predictions will be made only for the vehicles driving completely on these lanes.}
    \label{fig:lane_numeration}
\end{figure}

No prediction will be made for the vehicles standing completely or partially on on- and off-ramps. This is done to simplify the conversion of the coordinate system since on- and off-ramps would often require an ad hoc conversion due to the fact that they do not run consistently parallel to the other lanes. A reference path is then needed. A logical choice of a reference path for each road would be the line dividing the two driving direction. Unfortunately, the coordinates of this line are not directly included in the datasets under study. To produce an approximation of this line, a support vector machine (SVM) with a non-linear kernel (radial basis function) is applied to divide the scattered trajectory points relative to the vehicles driving in the two most internal lanes per driving direction. For example, the result of the SVM method for road 1 in the exiD dataset is shown in Fig. \ref{fig:SVM}. In this case the SVM was applied to lanes 3 and 5 (the lane numeration for the shown segment of road 1 is presented in Fig. \ref{fig:lane_numeration}). As the path has to follow the direction of travel, opposite directions of travel will have an inverted path.

The resulting boundary is not ready to be used as a reference path for the Cartesian-Frenet conversion as at the current state it is only a collection of points in Cartesian coordinates $p_i=(x_i,y_i)$. The tangent angle $\theta_i$ is needed for each point of the boundary $p_i$ and can be calculated as:
\begin{equation}
    \theta_i=\textnormal{arctan}(\frac{\dot y_i}{\dot x_i})
\end{equation}
Since the function of the boundary is not directly available, the approximated angle $\tilde{\theta}_i$ will be used instead and it can be calculated as:

\begin{equation}
    \tilde{\theta}_i = \textnormal{arctan}(\frac{\partial {y}_i}{\partial x_i}) 
\end{equation}

with:
\begin{gather}
    \partial x_i=x_{i+1}-x_{i-1}\\
    \partial y_i=y_{i+1}-y_{i-1}
\end{gather}
Now that the reference path is ready and it is a sequence of points $p_i=(x_i, y_i, \tilde{\theta}_i)$ with $i=1,2,...$, the Cartesian-Frenet conversion can be performed. After identifying the closest reference point $p_r=(x_r, y_r, \tilde{\theta}_r)$ to $p=(x,y)$, the following conversion formulas to find the corresponding Frenet coordinates can be applied:
\begin{gather}
    s=\sum^r_{i=2}\sqrt{(y_i-y_{i-1})^2+(x_i-x_{i-1})^2}\\
    |l|=\sqrt{(y-y_r)^2+(x-x_r)^2}\\
    l=\textnormal{sign}((y-y_r)\textnormal{cos}\tilde{\theta}_r-(x-x_r)\textnormal{sin}\tilde{\theta}_r)|l|
\end{gather}
When $r=1$ then $s=0$ . It should be noted that, given the way the reference path was designed, all the trajectory points end up with $l<0$.

To avoid those situations in which the road geometry would affect the lane change behavior (which would require a specific prediction algorithm), the curvature of the reference path is taken in consideration. 

The curvature $k_i$ for each point of the reference path $p_i=(x_i,y_i,\tilde{\theta}_i)$ can be calculated as:
\begin{equation}
    k_i = \frac{\dot x_i \ddot y_i-\dot y_i \ddot x_i}{(\dot x^2_i+\dot y^2_i)^{\frac{3}{2}}}
\end{equation}
Again, as the function of the boundary is not directly available, the approximated curvature  $\tilde{k}_i$ will be used instead and it can be calculated as:

\begin{equation}
    \tilde{k}_i=\frac{\partial x_i \partial^2 y_i-\partial y_i \partial^2 x_i}{(\partial  x^2_i+\partial y^2_i)^{\frac{3}{2}}}
\end{equation}

with:
\begin{gather}
    \partial^2 x_i=\partial x_{i+1} - \partial x_{i-1}\\
    \partial^2 y_i=\partial y_{i+1} - \partial y_{i-1}
\end{gather}
The resulting approximated curvature $\tilde{k}_i$ tends to be noisy. A moving average of window size $16$ is applied to smooth it out. To limit the effect of road geometry on the driving behaviour, an upper limit of $0.001$ for the curvature is enforced and trajectory points $p$ whose closest reference point $p_r$ has a curvature $k_r>0.001$ are excluded from the frames of interest. The conversion from Cartesian to Frenet is now complete, the next step is to cut the samples and extract the features.

\subsection{Data cutting and labeling}
The adopted definition of LC instant is analogous to the one used in \cite{lc_pred}, i.e. the instant in which the vehicle center crosses the lane line. For each LC instant, a LC trajectory segment of length $\Delta t_o$ is identified. The prediction time $\Delta t_p$ related to each LC trajectory is extracted with a uniform distribution between $0\textnormal{s}$ and $\Delta t_{p,\textnormal{MAX}}$ (if not possible, the segment is discarded). Out of simplicity, no LC trajectory segment is selected that contains another LC instant. Depending on the direction of the LC following the segment, each segment is labeled either as a LLC or a RLC. Then, a single LK trajectory segment of length $\Delta t_o$ is selected for each trajectory when possible (if more than one segments are feasible, only one is chosen randomly).  As defined in \cite{lc_pred}, "a LK trajectory segment is defined as a trajectory segment which does not contain any LC instant and whose ending instant does not precede a LC instant by a time between $0\textnormal{s}$ and $\Delta t_{p,\textnormal{MAX}}$". All LK trajectory segments are labeled as LK.

For some trajectories, rapid lane changes and oscillations around the lane markings created problems when extracting LC input TS. In total, seven input TS were affected (six in the exiD dataset and one in the Hongkong dataset). For simplicity, these seven input TS were removed manually.

The remaining segments constitute the dataset used for training and testing. 

\subsection{Feature extraction}
A set of features will constitute the input of the transformer. In particular, the used features are the lateral and longitudinal positions of the target vehicle and the lateral and longitudinal distances of the surrounding vehicles with respect to the target vehicle.

For a trajectory point $p$, which corresponds to a single frame of an input trajectory, the longitudinal and lateral positions of the target vehicle are respectively the already calculated quantities $s$ and $l$.

For the calculations of the distances of the surrounding vehicles, the calculations for the left preceding vehicle will be shown exemplarily. For all the other vehicles the calculations are analogous. The calculations of the longitudinal and lateral distances of the left preceding vehicle at a generic trajectory point $p$ ($\Delta s_{lp}$ and $\Delta l_{lp}$ respectively) are:
\begin{gather}
    \Delta s_{lp} =  s_{lp}-s\\
    \Delta l_{lp} =  l_{lp}-l
\end{gather}
where $s_{lp}$ and $l_{lp}$ are respectively the longitudinal and lateral position of the left preceding vehicle at the frame correspondent to the trajectory point $p$.

Finally, each sample of the resulting dataset (which will later be used to train and test the networks) will be composed of an input multivariate time series, or trajectory sample, $\overline{X}$ and its label $\overline{y}$ defined as:
\begin{gather}
    \overline{X} \in \mathbb{R}^{n \times d}\\
    \overline{y} \in \{LK,LLC,RLC\}
\end{gather}
where $n=\Delta t_o f_{\textnormal{DS}}$  with $f_{\textnormal{DS}}$ being the frequency at which trajectories were recorded in the respective dataset and $d$ the number of features (18 in this case). Each row $\overline{x}_j$ of $\overline{X}$ is a vector $\overline{x}_j \in \mathbb{R}^d$ with $j = 1,...,n $ defined as:

\begin{gather}
    \overline{x}_j = [l, s, \Delta l_{p}, \Delta s_{p}, \Delta l_{f}, \Delta s_{f}, \Delta l_{lp}, \Delta s_{lp},   \notag\\ \Delta l_{la}, \Delta s_{la}, \Delta l_{lf}, \Delta s_{lf}, \Delta l_{rp}, \Delta s_{rp}, \Delta l_{ra},  \notag\\ \Delta s_{ra}, \Delta l_{rf}, \Delta s_{rf}]
\end{gather}

where the pedicels (p, f, lp etc.) indicate the surrounding vehicles (see Figure \ref{fig:scenario}). The $j^{th}$ time-step of the input trajectory $\overline{X}$ will be referred to as $\overline{x}_j$.

Two final considerations need to be made to tackle two big differences between the two datasets under consideration. The first big difference is that in Hong Kong the driving direction is inverted with respect to Germany (i.e. road users drive "on the left"). While this may seem like a big issue, thanks to how the data was processed during the Cartesian to Frenet coordinates conversion this difference was eliminated by flipping the driving directions (an assumption is made that the behaviors are specular when the driving direction is inverted). The second big difference is that the two datasets were recorded at different frequencies. As stated earlier, the exiD dataset was recorded at $25$Hz while the Hongkong dataset was recorded at $30$Hz. This issue was resolved by interpolating the Hongkong samples to reduce their length (from 60 to 50 frames in this example).

For each sample, the average of the longitudinal and lateral positions were calculated and subtracted from the actual values of the positions to try to reduce the effect of road geometry. All the inputs were subsequently normalized before being used to train the transformer networks. 

To keep the datasets balanced, the number of samples for class LLC and RLC were set to be the same and the one for class LK was set to be double that. Moreover, the number of samples per class was set to be the same between the exiD and the Hongkong datasets. When, after the processing of the data, the number of available sample per class was greater than the number selected to maintain balance in the datasets, the desired number of samples was extracted randomly.

\section{Methodology}\label{sec:methodology}
In this section the machine learning method chosen to solve the problem of interest is analyzed. An introduction to the general architecture is presented  followed by an overview of the specific configuration adopted in this study. 

The methodology of this article follows that of our previous work \cite{lc_pred}. We present it here again in a more compact form to not hinder the readability of this work.

\begin{figure}
    \centering
    \begin{subfigure}{(a)}
        \centering
        \includegraphics[height=2.3in]{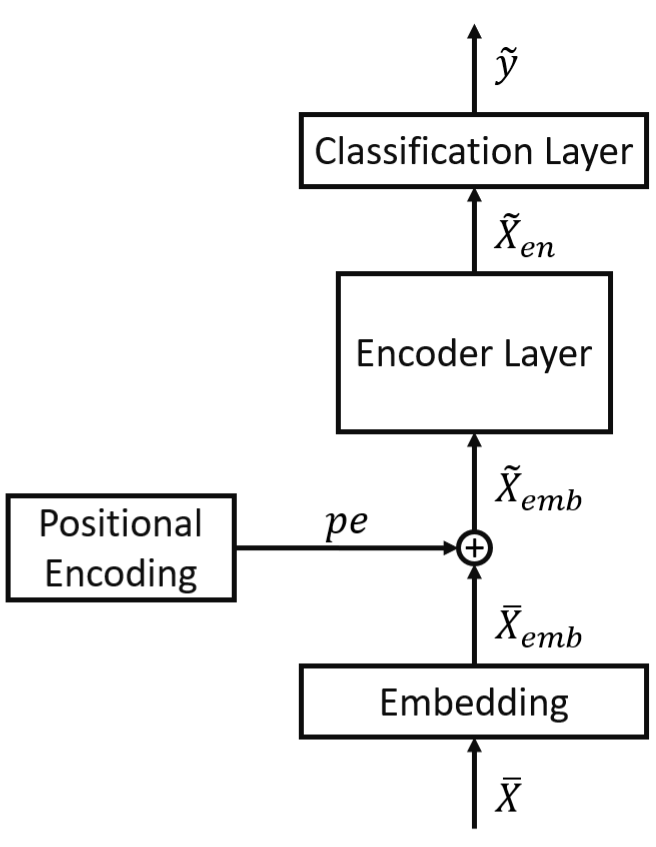}
    \end{subfigure}
    \begin{subfigure}{(b)}
        \centering
        \includegraphics[height=2.3in]{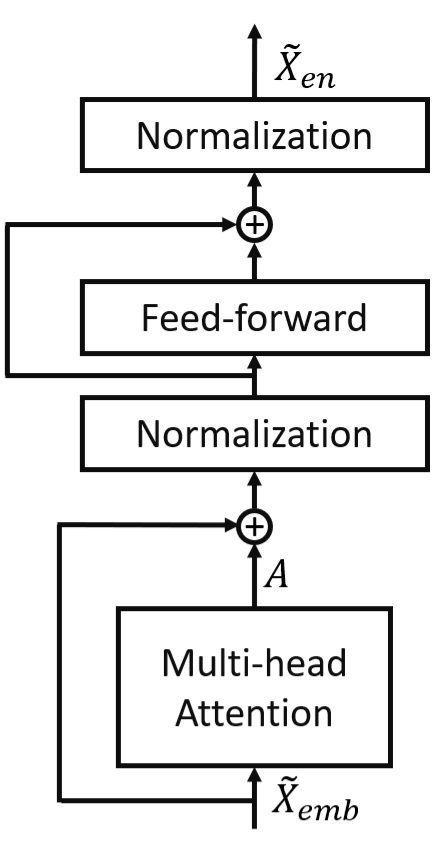}
    \end{subfigure}
    \caption{Structure of the transformer network (a) and close-up of the encoder layer (b). Figures taken from \cite{lc_pred}, licensed by CC BY 4.0 (see \cite{ccby4.0}).}
    \label{TN_structure}
\end{figure}
\begin{figure}
    \centering
    \includegraphics[width=0.4\linewidth]{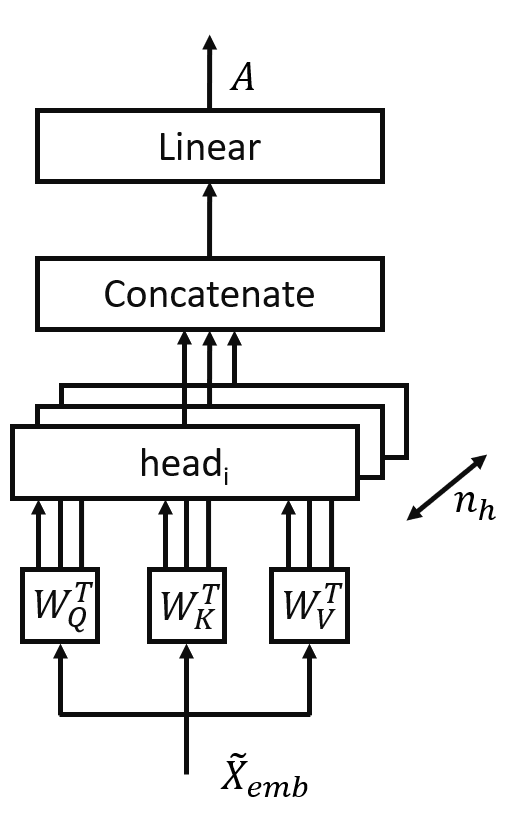}
    \caption{Structure of the multi-head attention layer. Figure taken from \cite{lc_pred}, licensed by CC BY 4.0 (see \cite{ccby4.0}).}
    \label{multi-head}
\end{figure}

A transformer network was selected for solving the lane change intention prediction problem under scope given its proven efficacy in similar situations. Transformer Networks (TNs), often referred to as Transformers, were introduced as a family of neural networks in 2017 by Vaswani et all. \cite{ref_trans}. The key idea of TNs is to find relationships between the values of an input time series and exploit them to generate an output. TNs typically have an encoder-decoder structure. Since in our case the problem to be solved is a classification one, only the encoder is employed while the decoder is substituted by a classification layer as shown in Fig. \ref{TN_structure}. The embedding layer is a linear function $f_{emb}$ that transforms an input multivariate time-series (a trajectory segment $\overline{X}$) into an embedded input multivariate time-series $\overline{X}_{emb}$:
\begin{equation}
    \overline{X}_{emb}=f_{emb}(\overline{X})
\end{equation}
with $\overline{X} \in \mathbb{R}^{n \times d}$ and $\overline{X}_{emb} \in \mathbb{R}^{n \times d_{emb}}$. The positional encoding's ($pe$) (which codifies the input time series' structure) is defined as:
\begin{equation}
    pe_{i,j}=\begin{cases}
        \textnormal{sin}((i-1)/1000^{(j-1)/d_{emb}})\;\;\textnormal{if\;\;j\;\;is\;\;odd}\\
        \textnormal{cos}((i-1)/1000^{(j-2)/d_{emb}})\;\,\textnormal{if\;\;j\;\;is\;\;even}
    \end{cases}
\end{equation}
with $i = 1,...,n$, $j = 1,...,d_{emb}$ and $pe \in \mathbb{R}^{n \times d_{emb}}$. The positional encoding is then added with a dropout rate of $0.1$ to $\overline{X}_{emb}$ and the resulting $\tilde{X}_{emb}\in\mathbb{R}^{n \times d_{emb}}$ is calculated as:
\begin{equation}
    \tilde{X}_{emb}=\overline{X}_{emb}+pe
\end{equation}
which is then passed to the encoder layer shown in Fig. \ref{TN_structure}. The structure of the multi-head attention block of the encoder layer is shown in Fig. \ref{multi-head}. At first, $\tilde{X}_{emb}$ is projected into query, key and value matrices $Q$, $K$, $V\in\mathbb{R}^{n \times d_{emb}}$:
\begin{gather}
    Q=\tilde{X}_{emb}W_Q^T\;,\;\;W_Q\in \mathbb{R}^{d_{emb} \times d_{emb}}\\
    K=\tilde{X}_{emb}W_K^T\;,\;\;W_K\in \mathbb{R}^{d_{emb} \times d_{emb}}\\
    V=\tilde{X}_{emb}W_V^T\;,\;\;W_V\in \mathbb{R}^{d_{emb} \times d_{emb}}
\end{gather}
For each head $i=1,...,n_h$ of the multi-head attention block the relative attention $a_i$ is computed as:
\begin{gather}
    a_i=softmax(\frac{QW_{q,i}^T(KW_{k,i}^T)^T}{\sqrt{d_h}})VW_{v,i}^T
\end{gather}
where $W_{q,i}$, $W_{k,i}$, $W_{v,i}\in\mathbb{R}^{d_h \times d_{emb}}$ and $d_h=\lfloor d_{emb}/n_h\rfloor$ for $i=1,...,(n_h-1)$. For $i=n_h$ instead, $W_{q,i}$, $W_{k,i}$, $W_{v,i}\in\mathbb{R}^{d_r \times d_{emb}}$ and $d_r=d_{emb}-n_hd_h$. The resulting attentions are then concatenated and linearly combined to generate the multi-head attention $A$:
\begin{equation}
    A=\begin{bmatrix}
        a_1 & ... & a_{n_h}
    \end{bmatrix}W_A^T
\end{equation}
where $W_A$ is a weight matrix with $W_A\in\mathbb{R}^{d_{emb} \times d_{emb}}$ and $A\in\mathbb{R}^{n \times d_{emb}}$. The output $\overline{X}_{en}\in\mathbb{R}^{n \times d_{emb}}$ of the encoder layer is then computed as:
\begin{equation}
    \overline{X}_{en}=Norm(Norm(A+\tilde{X}_{emb})+FF(Norm(A+\tilde{X}_{emb})))
\end{equation}
where $Norm()$ indicates a normalization layer and $FF()$ indicates a feed-forward layer of width $w_{FF}$. $\overline{X}_{en}$ needs now to be reduced to a single vector $\tilde{X}_{en}\in\mathbb{R}^{d_{emb}}$. In order to do so, for each of its column the average is computed:

\begin{gather}
   \overline{X}_{en} =
   \begin{bmatrix}\hat{x}_{1,1} & ... & \hat{x}_{1,d_{emb}}\\ \nonumber
    ... & \ddots & ... \\
    \hat{x}_{n,1} & ... & \hat{x}_{n,d_{emb}}
    \end{bmatrix} \in\mathbb{R}^{n \times d_{emb}} \\
    \Downarrow \\ \nonumber
    \tilde{X}_{en} = \begin{bmatrix} \frac{1}{n}\sum^n_{i=1}\hat{x}_{i,1} & ... & \frac{1}{n}\sum^n_{i=1}\hat{x}_{i,d_{emb}}
    \end{bmatrix} \in\mathbb{R}^{d_{emb}}
\end{gather}

Finally, $\tilde{X}_{en}$ passes through a linear classification layer which outputs $\tilde y$ which is a vector containing three values, one per class. Each sample is assigned to the class for which the respective output value is the highest among the three output values.

The configuration used in this work is identical to the configuration of TN 2 in \cite{lc_pred} i.e. it has a single encoder layer, $16$ multi-head attention heads and is optimized with \textit{Adam} \cite{ref_adam}. The transformer used in this article will also therefore be referred to as TN 2. The dimension of the embedding and the width of the feed forward layer are also identical (respectively $128$ and $64$) but the learning rate was reduced to $0.0004$ to reduce oscillations in the optimization process which were observed with a learning rate of $0.0007$.

\section{Results}\label{sec:results}
The evaluation metrics used in this article are accuracy and $F_1$ score which are standard for classification problems. Accuracy is the number of correct predictions over total number of predictions, $F_1$ score is a class-specific evaluation metric which is calculated as the harmonic mean between precision (true positives over true and false positives for a specific class) and recall (true positives over true positives and false negatives for a specific class). A detailed definition of these two metrics can be found in \cite{lc_pred}. To test the possibility of training a transformer on a population A and deploying it in a different population B for the purpose of LC intention prediction, two transformers (of the type described in section \ref{sec:methodology}) were trained: one on exiD data, one on Hongkong data. Both datasets were divided between a training dataset ($80\%$ of the data) and a testing dataset ($20\%$ of the data). Then, the two transformers were tested on the exiD and Hongkong datasets. The results are shown in Tab. \ref{tab:trans_results}: it is clearly observable that when a transformer is trained and tested on the same dataset A the performances are better (higher accuracy and $F_1$ scores) compared to those cases in which a transformer was trained on a different dataset B and tested on A despite the similarity of the scenarios.

\begin{table}[]
    \centering
    \begin{tabular}{c||c|c||c|c}
        Train data & \multicolumn{2}{c||}{exiD} & \multicolumn{2}{c}{Hongkong}   \\ \hline
        Test data & exiD & Hongkong & exiD  & Hongkong \\ \hline
        Acc. &$85.44\%$&$71.94\%$&$81.96\%$&$75.63\%$ \\
        $F_{1,LK}$ &$85.98\%$&$73.68\%$&$83.45\%$&$78.83\%$ \\
        $F_{1,LLC}$ &$85.58\%$&$68.21\%$&$80.51\%$&$72.92\%$ \\
        $F_{1,RLC}$ &$84.25\%$&$71.70\%$&$79.80\%$&$69.92\%$

    \end{tabular}
    \caption{Prediction results of the designed transformer for different combinations of training and testing datasets.}
    \label{tab:trans_results}
\end{table}

A possible explanation to this delta in the results when testing and cross-testing TN 2 is that there may be differences in the traffic conditions previously ignored. Longitudinal velocities were not included earlier among input features but might help characterize better traffic conditions. An approximation of the longitudinal velocities $\tilde{\dot{s}}_i$ for each time instant $i$ of an input trajectory can be calculated as:

\begin{equation}
    \tilde{\dot{s}}_i = \frac{\partial s_i}{2\delta t}
\end{equation}

where $\partial s_i = s_{i+1}-s_{i-1}$ and $\delta t=0.04$s, i.e. the duration of a time instanti in the exiD dataset and the interpolated Hongkong dataset. Of particular interest is the average longitudinal velocity of an input sample. The average longitudinal velocity for an input sample is calculated as the arithmetic mean of the approximated longitudinal velocity along the whole input sample.

\begin{figure}
    \centering
    \includegraphics[width=1\linewidth]{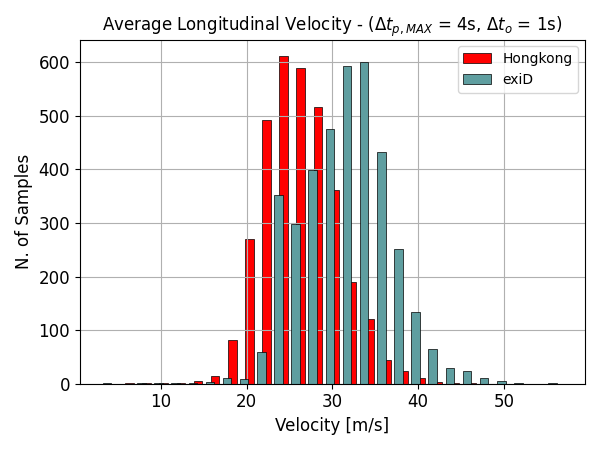}
    \caption{Distribution of the average longitudinal velocities for the lane change samples.}
    \label{mean_xVel_LLC}
\end{figure}

Looking at the distributions of the average longitudinal velocities of the LC samples in the processed datasets for $\Delta t_o=1$s in Figures \ref{mean_xVel_LLC}, it appears that German samples present, on average, a higher average longitudinal velocity with respect to the Chinese ones. To observe if these differences in average longitudinal velocities and in behavior in traffic were the cause of the poor performances observed earlier, two TN 2 were trained again on the exiD and the Hongkong datasets excluding the samples having an average longitudinal velocity lower than $20$m/s and higher than $30$m/s. These values were chosen because the distributions overlap in the interval $[20\textnormal{m/s},30\textnormal{m/s}]$ and training exclusively in this interval would mean that only samples extracted from similar traffic scenarios would be considered. The number of samples per class was again re-balanced and again divided in $80\%$ samples for training and $20\%$ for testing. The results of testing and cross-testing TN 2 on this "velocity capped" data are shown in Table \ref{tab:trans_results_cap}. It is evident that the differences in performances are still present: transformers trained and tested on a particular dataset perform clearly better than those trained on a different one, even if samples have now very similar average longitudinal velocities.

\begin{table}[]
    \centering
    \begin{tabular}{c||c|c||c|c}
        Train data & \multicolumn{2}{c||}{exiD} & \multicolumn{2}{c}{Hongkong}   \\ \hline
        Test data & exiD & Hongkong & exiD  & Hongkong \\ \hline
        Acc. &$86.84\%$&$71.58\%$&$79.47\%$&$72.37\%$ \\
        $F_{1,LK}$ &$87.18\%$&$73.52\%$&$80.31\%$&$74.76\%$ \\
        $F_{1,LLC}$ &$83.80\%$&$70.59\%$&$76.30\%$&$67.88\%$ \\
        $F_{1,RLC}$ &$89.01\%$&$68.66\%$&$80.61\%$&$71.04\%$ 
         
    \end{tabular}
    \caption{Prediction results of the designed transformer for different combinations of training and testing datasets with only samples with an average longitudinal velocity comprised between $20$m/s and $30$m/s.}
    \label{tab:trans_results_cap}
\end{table}

Finally, a transformer was trained on both the exiD and the Hongkong datasets. This transformer was then tested on the exiD and the Hongkong datasets separately. The results are shown in Tab. \ref{tab:trans_results_mixed}. This transformer showed good results for both the datasets on which it was tested on. The results on both are comparable to those previously obtained by the transformers trained and tested on the same datasets (Tab. \ref{tab:trans_results}).

\begin{table}[]
    \centering
    \begin{tabular}{c||c|c}
        Train data & \multicolumn{2}{c}{exiD + Hongkong}  \\ \hline
        Test data & exiD & Hongkong  \\ \hline
        Acc. &$86.29\%$& $75.95\%$ \\
        $F_{1,LK}$ &$87.01\%$& $78.12\%$ \\
        $F_{1,LLC}$ &$84.87\%$& $74.53\%$ \\
        $F_{1,RLC}$ &$86.18\%$& $72.14\%$
         
    \end{tabular}
    \caption{Prediction results of the designed transformer when using combined exiD and Hongkong datasets for training.}
    \label{tab:trans_results_mixed}
\end{table}

\section{Discussion}\label{sec:discussion}
The results presented in Tab. \ref{tab:trans_results} suggest that training transformer network on a population and testing it on a different one results in limited performances. Even trying to reduce the effect of the different traffic situations by only considering samples with similar average longitudinal velocity (see Tab. \ref{tab:trans_results_cap}) does not improve the results, suggesting that the cause of the poor performances must be different. This is of interest for manufacturers, as testing of prediction modules in a country seem to not guarantee how well the prediction module will perform in a different one. Even more so, it seems to suggest that the preferred approach would be to deploy specialized prediction modules for each region, since a transformer trained and tested on the same population shows instead significantly higher accuracy and $F_1$ scores. 

Although functional, this solution would present new obstacles: multiple transformers, trained on different populations, would be needed and a system would need to be implemented to correctly select the transformer that works the best in the region in which the end user is driving. A solution to these issues could be represented by a transformer trained on a mix of multiple populations. In this article, one transformer was trained on a mix of exiD and Hongkong datasets and it shows results, shown in Tab. \ref{tab:trans_results_mixed}, as good as those of the transformers trained and tested on the same population, both for exiD dataset and Hongkong dataset. This suggests that a transformer trained on multiple populations can perform as good on a single population as a transformer trained solely on that population.

A second observation can be made on the results shown in Tab. \ref{tab:trans_results} and Tab. \ref{tab:trans_results_mixed}, i.e. a difference in the results was observed between the populations: the transformer trained on both the exiD dataset and the Hongkong dataset performed significantly better when tested on the exiD dataset than when tested on the Hongkong dataset. This was true also for the transformers trained and tested on the same dataset: the transformer trained and tested on the exiD dataset performed better than the one trained and tested on the Hongkong dataset. This could mean that possibly Chinese naturalistic trajectories are harder to predict than German ones or that the architecture of the transformer, which was originally optimized on the highD dataset (German) in \cite{lc_pred}, needs to be optimized differently depending on the population on which it is trained and tested on. Given the limited amount of data of this study no final conclusion could be made without doubt.

\section{Conclusion}\label{sec:conclusion}
In the study presented in this article we tested the training a transformer on a population to use it to make prediction on a different one. Our results show that this is possible but with serious limitations and that on the other hand by training on both populations the transformer is able to achieve good performances on both. The results obtained on the German data were also significantly better than those obtained on the Chinese data, suggesting possibly that Chinese maneuvers are harder to predict or that different architectures work better with different populations. Future research should test these conclusions on a greater and more varied amount of data, which could give definitive answers to the issues that we found with our experiments. In addition, further investigations are needed to highlight if differences in the driving style or if differences in the scenarios are the cause of the lack in performances of the transformers trained on a population and tested on a different one.

\section*{Acknowledgments}
The research leading to these results has received funding from the Republic of Austria, Ministry of Climate Action, Environment, Energy, Mobility, Innovation and Technology through grant Nr.~891143 (TRIDENT) managed by the Austrian Research Promotion Agency (FFG). We would like to thank the Science Technology Plan Project of Zhejiang Province (Project Number: 2022C04023) and the Zhejiang Asia-Pacific Intelligent Connected Vehicle Innovation Center Co., Ltd. This work was partially supported by them. In addition, we would also like to thank leveLXData for providing us with useful datasets. A Large Language Model (ChatGPT, OpenAI) was used to assist us, the authors, with writing parts of the Python code which was used to produce the results and figures published in this work.

\newpage

\vfill

\end{document}